\RequirePackage{amsmath}
\documentclass[smallextended]{svjour3}       % onecolumn (second format)
\smartqed  % flush right qed marks, e.g. at end of proof
\journalname{Machine Learning}
\usepackage{graphics,graphicx,epsfig}
\usepackage{algorithm,algorithmic}
\usepackage{multirow,bm}
\usepackage{mathrsfs,amssymb}
\usepackage{natbib}
\usepackage{lmodern}% http://ctan.org/pkg/lm
\DeclareMathOperator{\argmax}{\arg\max}
\newcommand{\pr}[1]{\Pr\big(#1\big)}
\newcommand{\qw}[1]{\mathbb{E}\big[#1\big]}
\newcommand{\gongshi}[1]{equation \ref{#1}}
\newcommand{\citepsec}[1]{section \ref{#1}}

\newcommand{\citeplem}[1]{lemma \ref{#1}}

\newcommand{\intinf}[1]{\mathrm{d}#1}
\begin{document}
\title{Crowdsourcing with Unsure Option}

\author{Yao-Xiang Ding\and
        Zhi-Hua Zhou%etc.
}


\institute{Yao-Xiang Ding \at
  National Key Laboratory for Novel Software Technology\\
Nanjing University, Nanjing 210093, China\\
              \email{dingyx@lamda.nju.edu.cn}           %  \\
           \and
           Zhi-Hua Zhou \at
  National Key Laboratory for Novel Software Technology\\
Nanjing University, Nanjing 210093, China\\
              \email{zhouzh@lamda.nju.edu.cn}\\
              Corresponding author
}

\date{Received: date / Accepted: date}

\maketitle

\begin{abstract}
One of the fundamental problems in crowdsourcing is the trade-off between the number of the workers needed for high-accuracy aggregation and the budget to pay. For saving budget, it is important to ensure high quality of the crowd-sourced labels, hence the total cost on label collection will be reduced. Since the self-confidence of the workers often has a close relationship with their abilities, a possible way for quality control is to request the workers to return the labels only when they feel confident, by means of providing unsure option to them. On the other hand, allowing workers to choose unsure option also leads to the potential danger of budget waste. In this work, we propose the analysis towards understanding when providing the unsure option indeed leads to significant cost reduction, as well as how the confidence threshold is set. We also propose an online mechanism, which is alternative for threshold selection when the estimation of the crowd ability distribution is difficult.

\end{abstract}

\section{Introduction}
\label{Intro}
Labeled data play a crucial role in machine learning. In recent years, crowdsourcing has been a popular cost-saving way for label collection. The power of crowdsourcing relies on two conditions. One is the possibility to obtain highly accurate estimation of true labels by aggregating the collected noisy labels. Another is that the cost paid to the workers during the label collection process is not large, hence crowdsourcing is much more economical than to recruit domain experts. Particularly, in practice the budget to pay is usually limited. So it is important to study the approaches for balancing between cost reduction and estimation performance.

Unfortunately, there is a trade-off between aggregation accuracy and cost. As more labels are collected, typically, the aggregated accuracy increases, while the cost also increases. One way to deal with this problem is to design better label aggregation methods, without controlling the data collection process \citep{raykar2010learning, dalvi2013aggregating, zhang2014spectral}. Another more active way is to design effective task assignment mechanisms, for saving budget meanwhile maintaining the aggregation quality \citep{karger2011iterative, ho2013adaptive}. However, all these methods do not utilize the subjective behavior of the workers. Though seldom studied previously, it is interesting to consider an alternative kind of mechanisms utilizing the subjective uncertainty of the crowd, by allowing workers to choose unsure option instead of actually labeling the data. The advantage is that since the confidence of the workers often has a close relationship with their potential abilities, the quality of the returned labels may be improved. In \citep{zhongactive}, the setting of providing unsure option was studied under the active learning with crowd scenario, and the effect of label quality improvement was empirically justified from experiments. On the other hand, for ensuring the honesty of one worker, choosing unsure option should also be paid. Otherwise the workers would prefer to make guesses when their confidence is low. As a result, providing unsure option also leads to the potential danger of budget waste, since under the same budget, the number of returned labels is decreased. It is important to theoretically answer when providing unsure option can lead to significant cost reduction.

In this work, we take the first step towards the analysis of the cost-saving effect for the crowdsourcing with unsure option setting. Firstly, we provide the sufficient conditions for employing unsure option to be indeed effective on cost reduction. Secondly, we show how confidence threshold can be set properly. Thirdly, motivated by the theoretical results, we propose an alternative online mechanism. It is suitable to use for threshold selection when the statistics about the crowd are difficult to estimate due to the lack of golden standard tasks with known labels. 
% We show that simple statistics, i.e. mean and variance of the crowd ability, are key quantities in the analysis. We show that: (1) They are sufficient statistics to provide the sufficient condition on whether some unsure mechanism can indeed lead to significant cost reduction. (2) Based on them, the proper confidence threshold can be calculated. (3) When they can not easily be estimated due to the lack of the golden standard tasks with known labels, an online bandit based algorithm can be applied to set the confidence threshold. 

The rest of the paper is organized as follows. In section 2, the related work is discussed. In section 3, we describe the basic formulations and assumptions. Section 4 and section 5 give the theoretical analysis on the quality ensured and the unsure mechanisms. They include the main results of this paper. Section 6 discusses the possible extensions of the incentive compatible payment schemes. Section 7 introduces the online algorithm. Section 8 shows the experimental results. Section 9 concludes the paper.

\section{Related Work}
\label{RW}
Crowdsourcing, typically crowd-sourced data labeling, has been a fruitful topic in machine learning. One of the central tasks is to achieve desirable learning performance using the noisy labels returned by the crowd. This can be done by learning good classifiers utilising the noisy labels directly \citep{dekel2009good, dekel2009vox, urner2012learning}. Meanwhile, in many researches the label estimation problem has been thouroughly studied \citep{raykar2010learning, zhou2012learning, dalvi2013aggregating, li2014error, zhang2014spectral, zhou2015regularized}. The focus of these two kinds of researches is on improving learning performance, while the label collection cost is not directly considered. 

\citep{wang2016cost} points out the importance of the cost-saving effect in crowdsourcing, showing that cost reduction is one of the central tasks involved. And \citep{wang2015crowdsourcing} theoretically shows that it is indeed possible to achieve desired accuracy with reasonable cost. From the algorithmic perspective, there are many researches about task assignment and budget allocation, which try to balance between aggregation accuracy and data collection cost. Both non-adaptive task assignment mechanisms \citep{karger2011iterative, tran2013efficient}, which assign the tasks off-line before the worker comes, and the adaptive mechanisms which assign the tasks on-line during the labeling process \citep{ho2013adaptive, chen2013optimistic, abbasilarge}, have been studied thoroughly. The accuracy-cost trade-off is also the major issue discussed in this paper. The difference is that the task assignment and budget allocation mechanisms focus on improving the behavior of the task requesters, while for designing the unsure mechanisms, the target is to improve the behavior of the workers, by utilizing their own uncertainty about the tasks. There are also several studies on employing the bandit model into crowdsourcing. In \citep{abraham2013adaptive}, a special crowdsourcing problem called the bandit survey problem was considered. In \citep{jain2014incentive}, the bandit model was employed to deal with the task assignment problem. In \citep{zhou2014optimal}, the bandit arm identification problem was employed for worker selection. All these works have different settings to our work.

There are not many works studying the unsure mechanisms in crowdsourcing. \citep{zhongactive} considered providing unsure option under the active learning from crowd scenario. In their work, the purpose of allowing unsure option was to improve worker reliability. This is similar to the purpose of employing unsure mechanism in our work. The difference is that in their work, they empirically justified that providing unsure option to the crowd could make labeling quality improved in active learning by experiments. While in our work, instead of considering the active learning scenario, we consider the more general crowdsoured labeling task. Furthermore, our focus is on the theoretical analysis of the cost-saving effect of employing unsure option. In \citep{shah2014double}, a double or nothing incentive compatible mechanism was proposed, to make workers behave honestly based on their confidence. Their proposed mechanism is provable to avoid spammers from the crowd, under the assumption that every worker wants to maximize the expected payment. In their setting, employing unsure option is a way for ensuring high quality of the returned labels. The potential accuracy-cost trade-off is not considered. While in our work, the main focus is on the accuracy-cost trade-off, other than designing incentive compatible mechanisms.

\section{Problem Formulation} 
\label{PF}
We consider the tasks of collecting binary labels of $\{-1,1\}$. The well-known Dawid-Skene model \citep{dawid1979maximum, zhou2012learning, zhang2014spectral, karger2011iterative, zhou2015regularized, li2014error} under the binary classification case is adopted. Under the D-S model, the tasks are assumed to be homogeneous. The homogeneity means that the potential cost for different tasks are the same. As a result, in the rest of the paper, we focus on dealing with the cost for one single task.

We adopt the anonymous worker assumption introduced in \citep{karger2011iterative} for modeling the crowd. Let $\theta_i$ be the {\bf ability} of the $i$th worker, i.e. the probability that the returned label is correct. The process of choosing a number of anonymous workers is modeled as independent random draws $\theta_1, \theta_2, \theta_3 \dots \in [0,1]$ over the {\bf crowd ability distribution} $\theta_i \sim A$. Once drawn, the worker is asked to return the label or to choose unsure option based on the crowdsourcing mechanism applied, such as the quality ensured mechanism and unsure mechanism introduced below, or return the label directly if no mechanism is applied. The accuracy of the returned label is decided by the ability of the worker. We assume that after label collection, labels are aggregated by majority voting, i.e., to estimate the label based on the choices of the majority. In this paper, we assume that the mean $\mu$ of $A$ should be larger than $1/2$, i.e.
\begin{equation*} 
  \mu > 1/2.
\end{equation*}
The assumption is based on the well known result such that for majority voting, the estimated label does not converge to the true label when $\mu \leq 1/2$. From this result, if $\mu \leq 1/2$ then any crowdsourcing mechanism that makes the estimated label asymptotically correct can trivially leads to cost reduction. So we only consider the non-trivial case such that $\mu > 1/2$. We also adopt the following assumption:
\begin{equation}
  \pr{\theta < \mu-1/2} = 0.
  \label{assumpA}
\end{equation}
It states that we do not allow the workers with very low abilties which are far from the mean to exist. This assumption tightens our analysis. Indeed, these workers may correspond to the malicious workers who aim at attacking the crowdsourcing system, and should be excluded by any effective crodsourcing mechanism. Since malicious attacks prevention beyonds the scope of this paper, we just leave this as a preset assumption. Besides the above assumptions, we do not assume the crowd ability distribution $A$ to be any specific distributions.

We define the {\bf confidence} of the workers to be the subjective accuracy they believe to have, which is denoted by $c_1,c_2, c_3 \dots \in [1/2,1]$. The minimum value is $1/2$ since the weakest choice in one honest worker's mind is to make random guess. Similar to the crowd ability distribution $A$, we assume that the confidence for workers is independently drawn from the {\bf crowd confidence distribution} $c_i \sim C$. Obviously, there is a close relationship between one's ability and confidence. Thus modeling this relationship is necessary. In this paper we adopt a two-step analysis. At first, we can consider the mechanism, which simply filters out low quality labels based on the ability without using the information of the confidence. By this subroutine, we can further consider the setting of providing unsure option, by introducing reasonable assumptions on the relationship between workers' confidence and abilities.

As a consequence, we define a {\bf quality ensured mechanism} as the following process: given an ability threshold $T$, when one worker is drawn for labeling, the label is accepted only when the ability is above $T$. This mechanism is ideal since it assumes that once a worker is drawn, the ability can be also obtained. An {\bf unsure mechanism} is a surrogate for the idealized mechanism. When an unsure mechanism is employed, a confidence threshold $T$ is adopted. One worker is asked to do labeling only when the confidence $c \geq T$, otherwise he/she is asked to use the unsure option, for example, to return the label ``$0$'' which represents the unsureness. 

The {\bf budget} is defined as the total cost on the label collection. For simplicity of the analysis, we assume that returning one label and choosing unsure option once are both paid for $1$, i.e. the total cost equals to the total number of workers involved in the task. We Also assume that workers behave honestly according to their true abilities and confidence. Under the assumption that each worker aims at maximizing their payments, the honesty can be satisfied by adopting the incentive compatible payment mechanisms, which is discussed in section \ref{DPS}.

The goal of our analysis is to answer the question that, on which kinds of crowd ability distribution, an quality ensured mechanism or an unsure mechanism is provable to be {\bf effective}: 
\begin{definition}
  %Denote $m',m$ as the provable cost needed with/without an quality ensured mechanism or an unsure mechanism. For label aggregation being correct with probability at least $1-\delta$, an quality ensured mechanism or an unsure mechanism is provable to be effective if 
  For label aggregation being correct with high probability, denote $m$ as the provable cost needed for the simple aggregation without any mechanisms, and $m'$ as the provable cost needed when utilizing a quality ensured or an unsure mechanism. Then the utilized mechanism is effective if 
  \begin{equation*}
    m' \ll m. 
  %  \label{def1}
   \end{equation*} 
\end{definition}
Note that this is just a qualitative definition, and will be further quantified in definition 2 and 3. The effectiveness is defined for verifying the significance of cost reduction. 

In crowdsourcing, it is usually not possible to estimate the ability or the confidence of any individual worker, since the potential number of the tasks for an individual to accomplish is usually very limited. On the opposite, the statistical properties of the crowd ability distribution $A$ and the crowd confidence distribution $C$, are more practical to estimate. As a result, our analysis focuses on utilizing some simple statistics of $A$ and $C$, which are also much easier to estimate than to exactly model the distributions. Our target is to derive:\\
(1) How to properly set the ability and the confidence threshold. \\
(2) On what kind of $A$ and $C$, the quality ensured and the unsure mechanism can be effective.

The following two inequalities are important for the analysis:
\begin{lemma}[Hoeffding's inequality, \citep{shalev2014understanding} lemma B.6]
  For independent random variables \\
  $X_1,X_2,\dots,X_m$ bounded in $[a,b]$ with $\qw{X_i} =\mu_i, i=\{1,2,\dots,m\}$, the following inequality holds:
  \begin{equation}
    \pr{\frac{1}{m}\sum\limits_{i=1}^m\mu_i -\frac{1}{m}\sum\limits_{i=1}^m X_i > \epsilon} < \exp(-2m\epsilon^2/(b-a)^2).
  \label{hoeffding}
  \end{equation}
\end{lemma}
\begin{lemma}[Bernstein's inequality, \citep{shalev2014understanding} lemma B.9]
  For independent random variables\\
  $X_1,X_2,\dots,X_m$ bounded in $[-M,M]$ with $\qw{X_i} =0, i=\{1,2,\dots,m\}$ \\
  and variance $\sigma_1^2, \sigma_2^2,\dots,\sigma_m^2$, the following inequality holds:
  \begin{equation}
    %\pr{\sum\limits_{i=1}^m X_i > \epsilon} < \exp(-\frac{\epsilon^2/2}{m\sigma^2+M\epsilon/3}).
    \pr{\sum\limits_{i=1}^m X_i > \epsilon} < \exp(-\frac{\epsilon^2/2}{\sum\limits_{i=1}^m\sigma_i^2+M\epsilon/3}).
  \label{bernstein}    
  \end{equation}
\end{lemma}

\section{Analysis on the Quality Ensured Mechanisms}
\label{GA}
In this section, we focus on studying the quality ensured mechanisms. The analysis in this section forms the foundation to the analysis of the unsure mechanism in \citepsec{UA}.

To show that a quality ensured mechanism is effective, it is necessary to compare between the cost bounds with and without the mechanism. The first result is on the cost needed for using simple aggregation without any mechanisms:
\begin{lemma}
Let $\mu$ denote the mean of crowd ability distribution $A$, for simple majority voting aggregation under the settings introduced in \citepsec{PF}, the aggregated label is correct with probability at least $1-\delta$ if the total cost $m$ satisfies 
  \begin{equation}
    m \geq \frac{2[1+\frac{2}{3}(2\mu-1)]\log\frac{1}{\delta}}{(2\mu-1)^2}.
    \label{equaNormalBer}
 \end{equation}
\label{lemmaNormalBer}
\end{lemma}
\begin{proof}
  Given labels $a_1, a_2,\dots,a_m,\; a_i\in \{-1,1\}$ from workers, the majority voting rule is:
  \begin{equation*}
    \hat y = 1  \leftrightarrow  \sum\limits_{i=1}^m a_i >=0, \quad \hat y = -1  \leftrightarrow \sum\limits_{i=1}^m a_i <0.
  \end{equation*}
  The target is to bound $\pr{y\hat y <0} = \pr{y\sum\limits_{i=1}^m a_i <0 }$. Note that for $\forall i$, $\qw{ya_i} = \pr{ya_i=1} - \pr{ya_i=-1} = 2\theta_i-1.$ Then $\qw{\sum\limits_{i=1}^m ya_i} = \sum\limits_{i=1}^m (2\theta_i-1)$. 
%  By assumption, $\qw{\sum\limits_{i=1}^m \theta_i} = m\mu$, then 
  By assumption, $\qw{a_i} =\theta_i,\; \qw{\theta_i} = \mu$, then 
  \begin{equation*}
    \begin{aligned}
    \pr{y\sum\limits_{i=1}^m a_i <0} & = \pr{\qw{y\sum\limits_{i=1}^m a_i} - y\sum\limits_{i=1}^m a_i > \qw{y\sum\limits_{i=1}^m a_i}}\\
%      & =\pr{\qw{y\sum\limits_{i=1}^m a_i} - y\sum\limits_{i=1}^m a_i > \qw{y\sum\limits_{i=1}^m a_i}}\\
%    & = \pr{\sum\limits_{i=1}^m (2\theta_i-1) - y\sum\limits_{i=1}^m a_i > \sum\limits_{i=1}^m (2\theta_i-1)}\\
%    & = \pr{m(2\mu-1) - \sum\limits_{i=1}^m (2\theta_i-1) 
%    + \sum\limits_{i=1}^m (2\theta_i-1) - y\sum\limits_{i=1}^m a_i > m(2\mu-1) } \\
%    & = \Pr\Big(\qw {\sum\limits_{i=1}^m (2\theta_i-1) + y\sum\limits_{i=1}^m a_i} 
%    - (\sum\limits_{i=1}^m (2\theta_i-1) + y\sum\limits_{i=1}^m a_i) >  m(2\mu-1) \Big)\\
    & = \pr{\sum\limits_{i=1}^m (2\mu-1-ya_i) > m(2\mu-1)}.
  \end{aligned}
  \end{equation*}
 $2\mu-1-ya_i$ is a zero mean random variable bounded in $[-2,2]$, with variance 
  \begin{equation*} 
    \begin{aligned}
      \qw{(2\mu-1-ya_i)^2} & =\qw{(2\mu-1)^2-2(2\mu-1)ya_i+(ya_i)^2} \\
      & =(2\mu-1)^2+\qw{(ya_i)^2}-2(2\mu-1)\qw{ya_i} \\
    & =(2\mu-1)^2+1-2(2\mu-1)(2\theta_i-1)). \\
  \end{aligned}
  \end{equation*}

  $\frac{1}{m}\sum\limits_{i=1}^m[(2\mu-1)^2+1-2(2\mu-1)(2\theta_i-1)] \to 1-(2\mu-1)^2$ as $m \to \infty$, and $\lim\limits_{\mu \to \frac{1}{2}}\frac{1-(2\mu-1)^2}{2\mu-1} = \infty$. Thus
%, so the $\frac{1}{(2\mu-1)^2}$ order in proposition \ref{propHoef} is hard to be improved via Bernstein's bound. Instead, 
  we use a lemma to bound the variance:

  \begin{lemma}[\citep{Boucheron2013Concerntration}, Corollary 3.2]
    If $\sum_{i=1}^m X_i$ has the bounded difference property with constant $c$, then
    \begin{equation*}
      \mathrm{Var}(\sum_{i=1}^m X_i)^2 \leq \frac{1}{4} \sum_{i=1}^m c^2.
    \end{equation*}
  \end{lemma}
  From this we obtain that $\mathrm{Var}(\sum\limits_{i=1}^m(2\mu-1-ya_i)) \leq m$ since the bounded difference property with constant $2$ holds. 
  By equation \ref{bernstein}, 
  \begin{equation*}
    \begin{aligned}
    \pr{y\hat y <0 } & < \exp(-\frac{m^2(2\mu-1)^2/2}{\sum\limits_{i=1}^m\mathrm{Var}(2\mu-1-ya_i)+2m(2\mu-1)/3})\\
                     & = \exp(-\frac{m^2(2\mu-1)^2/2}{\mathrm{Var}(\sum\limits_{i=1}^m(2\mu-1-ya_i))+2m(2\mu-1)/3})\\
                     & \leq \exp(-\frac{m(2\mu-1)^2/2}{1+2(2\mu-1)/3}).\\
                   \end{aligned}
  \end{equation*}
Let $\delta= \exp(-\frac{m(2\mu-1)^2/2}{1+2(2\mu-1)/3})$, solving for $m$ gives the desired result.  
\qed
\end{proof}
The main order term in equation \ref{equaNormalBer} is $\frac{1}{(2\mu-1)^2}$. Then we show another lemma, which gives the cost bound for a quality ensured mechanism. 
\begin{lemma}
Let $T>1/2$ be the ability threshold, and $\eta=\pr{\theta \geq T}$ be the upper tail probability of the crowd ability distribution $A$. When the quality ensured mechanism is employed, for majority voting aggregation under the settings introduced in section \ref{PF}, the aggregated label is correct with probability at least $1-\delta$, if the total cost $m'$ satisfies 
  \begin{equation}
    m' \geq \frac{2(1-\eta)\log\frac{2}{\delta}}{\eta}+\frac{4\log\frac{2}{\delta}}{(2T-1)^2\eta}+\frac{2}{3\eta}.
    \label{equaUnsureBer}
 \end{equation}
  \label{lemmaUnsureBer}
\end{lemma}
\begin{proof}
  The first step: For those workers whose confidence is larger than $T$, we bound the number of workers needed when the aggregated estimation is incorrect with probability at most $\delta/2$, which is denoted as $m'_0$.

  The target is to bound $\pr{y\hat y <0} = \pr{y\sum\limits_{i=1}^{m'_0} a_i <0 }$. Note that for $\forall i$, $\qw {ya_i} = \pr{ya_i=1} - \pr{ya_i=-1} = 2\theta_i-1 \geq 2T-1.$ Then $\qw {\sum\limits_{i=1}^{m'_0} ya_i }\geq m'_0(2T-1)$. Then

 \begin{equation*}
    \begin{aligned}
      \pr{y\sum\limits_{i=1}^{m'_0} a_i <0 } & =\pr{\qw {y\sum\limits_{i=1}^{m'_0} a_i} - y\sum\limits_{i=1}^{m'_0} a_i > \qw {y\sum\limits_{i=1}^{m'_0} a_i }}\\
      & \leq \pr{\qw {y\sum\limits_{i=1}^{m'_0} a_i} - y\sum\limits_{i=1}^{m'_0} a_i > m'_0(2T-1)}.
     \end{aligned}
  \label{concernB1}
  \end{equation*}

  $ya_i$ is an independent random variable in $\{-1,1\}$, by equation \ref{hoeffding},

  \begin{equation*}
    \pr{y\hat y <0 } < \exp(-m(2T-1)^2/2).
    \label{concernB2}
  \end{equation*}
Let $\delta/2 = \exp(-m'_0(2T-1)^2/2)$, solving for $m'_0$ gives
\begin{equation}
  m'_0 = \frac{2\log(2/\delta)}{(2T-1)^2}.
  \label{concernB3}
\end{equation}

The second step: we bound the cost when sufficient number of workers have confidence larger than $T$.

Consider the random variable $\mathbb I(\theta>T)$, the target is to bound $\pr{\sum\limits_{i=1}^{m'}\mathbb I(\theta > T) < m'_0}$. We have
\begin{equation*}
  \begin{aligned}
  & \pr{\sum\limits_{i=1}^{m'}\mathbb I(\theta > T) < m'_0}\\
  & = \pr{\qw {\sum\limits_{i=1}^{m'}\mathbb I(\theta > T)} - \sum\limits_{i=1}^{m'}\mathbb I(\theta > T) > \qw {\sum\limits_{i=1}^{m'}\mathbb I(\theta > T)} - m'_0 }. \\
  \label{}
\end{aligned}
\end{equation*}
Since $\qw {\mathbb I(\theta>T)} = \pr{\theta>T}=\eta, \qw{\mathbb I(\theta>T)-\qw{ I(\theta>T)} }^2= \eta(1-\eta)$, by equation \ref{bernstein}, 
\begin{equation*}
  \pr{\sum\limits_{i=1}^{m'}\mathbb I(\theta > T) < m'_0} < \exp(-\frac{(m'\eta-m'_0)^2/2}{m'\eta(1-\eta)+(m'\eta-m'_0)/3}).
  \label{}
\end{equation*}

Let $\exp(-\frac{(m'\eta-m'_0)^2/2}{m'\eta(1-\eta)+(m'\eta-m'_0)/3}) = \delta/2$, we have
\begin{equation*}
  \frac{1}{2}\eta^2m'^2-[(1-\eta)\log\frac{2}{\delta}+m'_0+\frac{1}{3}]\eta m' + \frac{1}{2}{m'_0}^2+\frac{1}{3}m'_0\log\frac{2}{\delta} = 0,
\end{equation*}
then
\begin{equation*}
  m' = \frac{(1-\eta)\log\frac{2}{\delta}+m'_0+\frac{1}{3}+ \sqrt{[(1-\eta)\log\frac{2}{\delta}+m'_0+\frac{1}{3}]^2-{m'_0}^2-\frac{2}{3}m'_0\log\frac{2}{\delta}}}{\eta}.
  \label{}
\end{equation*}

It is easy to see that $m'=\frac{2((1-\eta)\log\frac{2}{\delta}+m'_0+\frac{1}{3})}{\eta}$ also satisfies the condition. Together with equation \ref{concernB3}, the desired result can be shown by union bound.
\qed
\end{proof}

The main order term in \gongshi{equaUnsureBer} is $\frac{1}{(2T-1)^2\eta}$. Given the above two cost bounds, we propose a more concrete definition on the effectiveness, based on how much reduction on the main order of the cost:
\begin{definition}
Let $\mu$ be the mean of the crowd ability distribution, $T$ be the ability threshold for a quality ensured mechanism and $\eta = \pr{\theta \geq T}$ be the probability for a worker to have the ability above $T$, then a quality ensured mechanism is at least {\bf $\alpha$-effective} if 
%  it is provable that for the aggregated label being correct with probability at least $1-\delta$,
  \begin{equation}
  \frac{1}{(2T-1)^2\eta} \leq \frac{1}{(2\mu-1)^\alpha}.
    \label{equadef2}
  \end{equation}
  \label{def2}
\end{definition}
When equation \ref{equadef2} is satisfied, by \citeplem{lemmaNormalBer} and \citeplem{lemmaUnsureBer}, the main order term of the bound improves in the order of $\frac{1}{2\mu-1}$, from $2$ to $\alpha$. $\alpha$ is a measure of significance, as $\alpha$ decreases, the improvement on the cost becomes more significant. We can also see that the ability threshold $T$ should be larger than the mean ability $\mu$, i.e.
\begin{equation*}
  T > \mu.
\end{equation*}
The reason is that from equation \ref{equadef2}, it is not possible for a quality ensured mechanism with $T\leq \mu$ to be $\alpha$-effective with $\alpha < 2$, which means that the mechanism can not lead to cost reduction.
%For the chosen ability threshold $T$, there exists a constant $\gamma$, such that
%\begin{equation}
%So we can assume that the following equation:
%  \begin{gather} 
%  T  = \mu + r, \; r > 0 \\
%  % \pr{\theta \geq \mu + r}  = \gamma \pr{\theta \leq \mu - r}.
%  \end{gather}
%\end{equation}
Now we are ready to show the main result of this section, a general sufficient condition for a quality ensured mechanism to be at least $\alpha$-effective. 
\begin{theorem}
  For crowd ability distribution with mean $\mu > 1/2$ and variance $\sigma^2$ under the condition in \gongshi{assumpA}, when 
  \begin{equation}
    (1+\gamma)\frac{(1+\sqrt{1-4\sigma^2})^2}{\sqrt{3}\sigma^2[2\sigma^2+(2\mu-1)]^2} \leq \frac{1}{(2\mu-1)^\alpha},
    \label{thm21}
  \end{equation}
  in which
  \begin{equation}
    \gamma = \frac{\pr{\theta \leq \mu - \frac{1}{2}\sqrt{1-\sqrt{1-4\sigma^2}}}}{\pr{\theta \geq \mu + \frac{1}{2}\sqrt{1-\sqrt{1-4\sigma^2}}}},
    \label{equaGamma}
  \end{equation}
  then the quality ensured mechanism with ability threshold
  \begin{equation}
    T= \mu + \frac{1}{2}\sqrt{1-\sqrt{1-4\sigma^2}}
    \label{thm22}
  \end{equation}
  is at least $\alpha$-effective.
  \label{theoremUnsureGeneral}
\end{theorem}
\begin{proof}
  First we show a lemma which gives a lower bound on the tail probability for the crowd ability distribution. The intuition is that when the variance of a random variable is high, the tail probability should not be too small.
\begin{lemma}
  Given a random variable $x \sim A, x\in [0,1]$ with mean $\mu$ and variance $\sigma^2$, under the condition in \gongshi{assumpA}, for $0 < r < \sigma$, we have
  \begin{equation*}
    \pr{|x-\mu|\geq r} \geq \frac{2\sqrt{3}(\sigma^2-r^2)}{1-4r^2}.
    \label{}
  \end{equation*}
\end{lemma}
\begin{proof}
  Suppose that 
  \begin{equation*}
    \pr{|x-\mu| > r} = A.
  \end{equation*}
  We can derive an upper bound of the variance $\sigma^2$. Denote $Y=|x-\mu|$, by $\mu > 1/2$ and \gongshi{assumpA}, we have $Y \in [0,1/2]$. Then 
\begin{equation*}
  \begin{aligned}
    \sigma^2 = \qw{ (x-\mu)^2} & =\int_0^{\frac{1}{2}} y^2\intinf{P(Y\leq y)}=\int_0^{\frac{1}{2}} 2y\pr{Y > y}\intinf{y}\\
    & = \int_0^{r}2y\pr{Y > y}\intinf{y} + \int_{r}^{\frac{1}{2}} 2y\pr{Y>y}\intinf{y}\\
    & \leq \int_0^{r}2y\intinf{y} + \sqrt{\int_{r}^{\frac{1}{2}}4y^2\intinf{y}\int_{r}^{\frac{1}{2}}(\pr{Y>y})^2\intinf{y}}\\
    & \leq \int_0^{r}2y\intinf{y} + \sqrt{\int_{r}^{\frac{1}{2}}4y^2\intinf{y}\int_{r}^{\frac{1}{2}}A^2\intinf{y}}\\        
    & \leq r^2 + \sqrt{[\frac{1}{6}-\frac{4}{3}r^3](\frac{1}{2}-r)A^2}.
  \end{aligned}
\end{equation*}
The first inequality is due to $\pr{Y>y} \leq 1$ and Cauchy-Schwarz inequality. The second inequality is due to $\pr{Y > y} \leq A$ when $y>r$. We then have
\begin{equation*}
  A \geq \frac{2\sqrt{3}(\sigma^2-r^2)}{\sqrt{[1-8r^3](1-2r)}}.
\end{equation*}
and 
\begin{equation*}
  \pr{|x-\mu|\geq r} \geq \pr{|x-\mu| > r} = A \geq \frac{2\sqrt{3}(\sigma^2-r^2)}{\sqrt{[1-8r^3](1-2r)}}.
\end{equation*}
Observe that 
\begin{equation*}
  (1-8r^3)(1-2r) = 1-((2r)^3+2r)+(2r)^4 \leq 1 - 2(2r)^2 + (2r)^4 = (1-(2r)^2)^2.
  \label{}
\end{equation*}
Then  
\begin{equation*}
  \pr{|x-\mu|\geq r} \geq \frac{2\sqrt{3}(\sigma^2-r^2)}{1-4r^2}.
  \label{equiLower1}
\end{equation*}
\hfill\qed
\end{proof}
Denote the ability threshold as $T = \mu + r > \mu$, and assume that 
\begin{equation*}
  \pr{\theta \leq \mu-r} = \gamma \pr{\theta \geq \mu + r},
\end{equation*}
then 
\begin{equation}
  \pr{\theta \geq T} = \pr{\theta \geq \mu + r} \geq \frac{1}{1+\gamma}\frac{\sqrt{3}(\sigma^2-r^2)}{1-4r^2}.
  \label{lowerBoundTransform}
\end{equation}
Now we turn to the task of finding the minimum of $\frac{1}{(2T-1)^2\eta}$, which is equivalent to maximizing $(2T-1)^2\eta$. Relaxation can be made to make use of the lower bound given by \gongshi{lowerBoundTransform}. Then we turn to the maximization of 
\begin{equation*}
  \frac{1}{1+\gamma}\frac{(\sqrt{3})(\sigma^2-r^2)}{1-4r^2}(2\mu+2r-1)^2.
  \label{}
\end{equation*}
We have a constraint $0<r<\sigma$, and this constraint is enough to ensure $T=\mu+r \geq 1$. To see this, for random variable $x$ in $[0,1]$ with mean $\mu_x > 1/2$ and variance $\sigma_x^2$, the maximum of the variance is attained on the distribution such that we have probabilities of $1/2$ only at $x=1$ and $2\mu-1$. Then we can verify that under this distribution, we have $\sigma^2_x = (1-\mu_x)^2$ and $\mu_x + \sigma_x = 1$.

By $0 < r < \sigma$, we have $(\frac{2\sigma+(2\mu-1)}{2\sigma})2r \leq 2r+2\mu-1$. We turn to the maximization of 
\begin{equation*}
  (4\sqrt{3})(\frac{1}{1+\gamma})(\frac{2\sigma+(2\mu-1)}{2\sigma})^2\frac{(\sigma^2-r^2)}{1-4r^2}r^2.
  \label{}
\end{equation*}
Let $s=r^2$ and drop the constants for the moment. We consider 
\begin{equation*}
 \max\limits_{0 < s < \sigma^2} f(s) = \frac{\sigma^2s-s^2}{1-4s}.
  \label{}
\end{equation*}
We then have 
\begin{equation*}
  f'(s) = \frac{(\sigma^2-2s)(1-4s)+4(\sigma^2s-s^2)}{(1-4s)^2}=\frac{4s^2-2s+\sigma^2}{(1-4s)^2}.
  \label{}
\end{equation*}
Without consideration of constraints on $s$, the optimal is attained when $s=(1-\sqrt{1-4\sigma^2})/4$. The solution satisfies the constraints since
\[ 1 - 4\sigma^2 \leq \sqrt{1-4\sigma^2} \iff \sigma^2 \geq (1-\sqrt{1-4\sigma^2})/4. \]
So we have  
\begin{equation*} 
  T= \mu + \frac{1}{2}\sqrt{1-\sqrt{1-4\sigma^2}}
  \label{}
\end{equation*}
and 
\begin{equation*}
  m' = (1+\gamma)\frac{(1+\sqrt{1-4\sigma^2})^2}{\sqrt{3}\sigma^2[2\sigma^2+(2\mu-1)]^2}.
  \label{}
\end{equation*}
\hfill\qed
\end{proof}

From the theorem, it can be seen that as $\sigma^2$ increases, the left hand side of \gongshi{thm21} decreases such that lower $\alpha$ can be achieved. In \gongshi{equaGamma}, the smaller the $\gamma$ is, the larger the upper tail probability at $T$ can be ensured. It is reasonable to assume that for an effective quality ensured mechanism, $\gamma$ can not be large, since $\pr{\theta \geq T}$ can not be small. 

In \gongshi{thm22}, $T$ increases as the variance gets larger. The intuition behind is that when the variance of the crowd ability distribution increases, then we have more workers with high ability, and we can safely increase $T$ to make higher demand on the quality of the returned labels.

\section{Analysis on the Unsure Mechanisms}
\label{UA}
In this section, we consider the unsure mechanisms. In the previous analysis, the quality ensured mechanisms guarantee that the abilities of the workers who return their labels are above the threshold $T$. However, due to the potential mismatch between one's confidence and ability, an unsure mechanism does not guarantee this property, since the workers behave based on their confidence, not their true abilities. The only reasonable assumption we can make is that there can be a positive correlation between the confidence and ability for an individual worker. This makes it difficult to estimate the mean ability of the crowd directly, which is essential for deriving the cost bound. In spite of this diffculty, one of the major theoretical findings in this section is: If we can filter out a bit more workers with low abilities besides the assumption in equation 1, then we can lower bound the mean ability. As a result, we introduce the following {\bf worker testing stage} conducted before the actual labeling tasks start: 
\begin{enumerate}
\item
Keep a small pool of golden standard tasks with known labels. 
\item
For each worker in the crowd, $k$ golden standard tasks are drawn for testing the ability. 
\item
We only send tasks to workers who correctly labels all $k$ golden standard tasks. 
\end{enumerate}
Note that we do not assume to have a sufficient number of golden standard tasks to accurately estimate each workers' abilities, or to make $k$ large so that the mean ability of workers can be boosted. On the opposite, we assume that $k$ is very small since by this it is enough to filter out a bit more low quality workers. The experimental results in section \ref{Exp} show that introducing the worker testing stage is effective even when $k=1$. We assume that no rewards are paid during the test stage, since the workers tend to have the motivation for passing the test. It is also essential to ensure that the workers behave the same among the test and the real tasks. As we assume workers' honesty in this paper, this is not a problem. While in applications it is necessary to utilize incentive compatible payment mechanism to ensure honesty, as discussed in section \ref{DPS}. 

% For an unsure mechanism, the workers behave based on their confidence. So we can assume that the confidence of workers is known by the mechanism. While the abilities are unknown information to the mechanism. For further analysis, modeling the relationship between the ability and the confidence is necessary. 
The next task is to model the relationship between one's ability and confidence. First we introduce some notations, as listed in table 1.
\begin{table}[h]
  \centering
  \begin{tabular}{|c|c|}
    \hline
    Notation & Meaning \\
    \hline
    $passed$ & The event that a worker passes the worker testing stage \\
    $A_{0}$ & Crowd ability distribution before the worker testing stage \\
    $A_{1}$ & Crowd ability distribution after the worker testing stage \\
    %$C_{0}$ & Crowd confidence distribution before the worker testing stage \\
    $C_{1}$ & Crowd confidence distribution after the worker testing stage \\
    $\mathrm{Pr}_{\theta,0}(\cdot), \mathrm{Pr}_{\theta,1}(\cdot)$ & Probability over $A_0, A_1$ \\
    %$\mathrm{Pr}_{c,0}(\cdot), \mathrm{Pr}_{c,1}(\cdot)$ & Probability over $C_0, C_1$ \\
    $\mathrm{Pr}_{c,1}(\cdot)$ & Probability over $C_1$ \\
    %$\mathbb{E}_{\theta,0}(\cdot), \mathbb{E}_{\theta,1}(\cdot)$ & Expectation over $A_0, A_1$ \\
    $\mathbb{E}_{\theta,0}(\cdot), \mathbb{E}_{c,1}(\cdot)$ & Expectation over $A_0, C_1$ \\
    %$\mathbb{E}_{c,0}(\cdot), \mathbb{E}_{c,1}(\cdot)$ & Expectation over $C_0, C_1$ \\
    $\mu_{\theta,0}, \sigma^2_{\theta,0}$ & Mean and variance of $A_0$ \\
    $\mu_{c,1}, \sigma^2_{c,1}$ & Mean and variance of $C_1$ \\
    $\eta_{\theta,0}, \eta_{\theta,1}$ & $\eta_{\theta,0} = \mathrm{Pr}_{\theta,0}(\theta \geq T)$, $\eta_{\theta,1} = \mathrm{Pr}_{\theta,1}(\theta \geq T)$\\  
    $\eta_{c,1}$ & $\eta_{c,1} = \mathrm{Pr}_{c,1}(c \geq T)$ \\
    \hline
  \end{tabular}
  \caption{Some notations used in further analysis.}
\end{table}
%Let $A_{\theta,0}, A_{\theta,1}$ be the crowd ability distribution before and after We use $passed$ to denote the event that a worker passes the worker testing stage. The mean and the variances of the crowd ability distribution $A$ before the worker testing stage are denoted as $\mu_{\theta,0}$ and $\sigma^2_{\theta,0}$. The mean of the crowd ability distribution after the worker testing stage is denoted as $\mu_{\theta,1}$. The mean and the variance of the crowd confidence distribution $C$ after the the worker testing stage are denoted as $\mu_{c,1}$ and $\sigma^2_{c,1}$. The chosen confidence threshold is denoted as $T$. $\eta_{\theta,0} = \pr{\theta \geq T}$, $\eta_{\theta,1} = \pr{\theta \geq T | passed}$ and $\eta_{c,1} = \pr{c \geq T | passed}$ are tail probabilities. Taking expectation with respect to the crowd ability distribution before the worker testing stage is denoted as $\mathbb E_{\theta,0}[\cdot]$. 

To follow the analysis process in \citepsec{GA}, we assume that 
\begin{equation}
  \begin{cases}
    T > \mu_{\theta,0}, & \quad \eta_{c,1} \geq \eta_{\theta,1}, \\
    T > \mu_{c,1},& \quad \eta_{c,1} < \eta_{\theta,1}. \\
  \end{cases}
\end{equation}
Furthermore, we introduce the following assumptions:\\
When $\eta_{c,1} \geq \eta_{\theta,1}$, there exists $k_{0} > 0$, for all $T > \mu_{\theta,0}$, 
  \begin{gather}
    % \pr{\theta > \mu_{\theta,0} + r}  \geq \pr{\theta < \mu_{\theta,0} - r}, \\
  \frac{\pr{c \geq T | \theta \geq T, passed}}{\eta_{c,1}} \geq \Big(\mathbb E_{\theta,0}[\theta^k]\Big)^{k_0}(\frac{1}{\eta_{\theta,0}}), \\
    \frac{(\mu_{\theta,0})^{k+1}}{(\mathbb E_{\theta,0}[\theta^k])^{1-k_0}} \geq 1/2.
  \end{gather}
  When $\eta_{c,1} < \eta_{\theta,1}$, then there exsits $k_1 \geq 0$, for all $T > \mu_{c,1}$,
%\begin{equation}
  \begin{gather}
    %\pr{c > \mu_{c,1} + r} \geq \pr{c < \mu_{c,1} - r}, \\
    \frac{\pr{c \geq T | \theta \geq T, passed}}{\eta_{c,1}} \geq (\mu_{c,1})^{k_1}(\frac{1}{\eta_{c,1}}),\\
    (\mu_{c,1})^{k_1+1} \geq 1/2.
    %\label{unsureassump2}
  \end{gather}
  Ignoring the exponential terms, equation 13 and 15 imply that 
  % when $k_0$ or $k_1$ is chosen sufficiently small, then 
  \[ \pr{c \geq T | \theta \geq T, passed} \geq \eta_{c,1} = \pr{c\geq T| passed}. \]
  This is a reasonable assumption, since for an unsure mechanism to be useful, the positive correlation between confidence and ability is necessary. The task dependent constants $k_0$ and $k_1$ control the magnitude of this positive correlation. Since we have $\mathbb E_{\theta,0}[\theta^k] < 1$ and $\mu_{c,1} < 1$, the larger $k_0$ and $k_1$ are, the weaker the positive correlation becomes. While these two constants are usually small since it is common for the confidence and the ability to be correlated. Equation 14 and 16 are adopted for cost bound derivation, ensuring the transformed threshold $T'$ to be above $1/2$ (See equation 18 and 19). Under the above assumptions, we can get the following cost bound for an unsure mechanism: 
%For further analysis, % And equation (20) is needed for ensuring that the average ability of workers who have confidence above $T$ to be larger than $1/2$ (see equation 23). 
\begin{lemma}
  Assume the conditions in equation (12-16) to hold. Employ the unsure mechanism with confidence threshold $T$ and the worker testing stage with $k$ golden standard tasks. For majority voting aggregation under the settings introduced in \citepsec{PF}, the aggregated label is correct with probability at least $1-\delta$ if the cost satisfies
    \begin{equation}
    m' \geq \frac{2(1-\eta)\log\frac{2}{\delta}}{\eta}+\frac{8\log\frac{2}{\delta}}{(2T'-1)^2\eta}+\frac{2}{3\eta}.
    \label{equaUnsureBer2}
    \end{equation}
    When $\eta_{c,1} \geq \eta_{\theta,1}$, 
 \begin{equation}
    \eta = 
    \Big(\frac{T^k}{\mathbb E_{\theta,0}[\theta^k]}\Big)\eta_{\theta,0}, \quad T' = \Big(\frac{T^k}{(\mathbb E_{\theta,0}[\theta^k])^{1-k_0}}\Big)T. 
  \label{}
\end{equation}
When $\eta_{c,1} < \eta_{\theta,1}$, 
 \begin{equation}
   \eta = \eta_{c,1}, \quad T' = (\mu_{c,1})^{k_1}T.
  \label{}
\end{equation}
\label{lemmaUnsureBer2}
\end{lemma}
%{\it Remark.} For the choice of $k$, when $\eta_{c,1} \geq \eta_{\theta,1}$, we can choose 
%\[k = \arg\max_{k} \frac{(\mu_{\theta,0})^{k+1}}{(\mathbb E_{\theta,0}[\theta^k])^{1-k_0}}.\]
%This makes the bound the tightest. And when $\eta_{c,1} < \eta_{\theta,1}$, the choice of $k$ does not affect the bound directly. Since larger $k$ makes more workers to be filtered out, in practice keeping $k$ small is desirable.
\begin{proof}
  The key idea is to estimate two quantities for the crowd after the worker testing stage. One is the proportion of the workers who have confidence above $T$, i.e. $\eta_{c,1}$. When $\eta_{c,1} < \eta_{\theta,1}$, we directly use $\eta_{c,1}$. Otherwise when $\eta_{c,1} \geq \eta_{\theta,1}$, we should lower bound $\eta_{\theta,1}$.
  % When $\eta_{c,1} \geq \eta_{\theta,1}$, we need to lower bound $\eta_{c,1}$. And in general we want to lower bound $\qw {\theta | c \geq T, passed}$.   
  \begin{equation*}
    \eta_{\theta,1} = \frac{\eta_{\theta,0}\pr{passed|\theta \geq T}}{\mathrm{Pr}_{\theta,0}(passed)}.
    \label{}
  \end{equation*}
  It is easy to see that $\pr{passed|\theta\geq T} \geq T^k$, and 
  \begin{equation*}
    \mathrm{Pr}_{\theta,0}(passed)= \mathbb E_{\theta,0}[passed|\theta]= \mathbb E_{\theta,0}[\theta^k].
    \label{}
  \end{equation*}
  So we have
  \begin{equation*}
  \eta_{\theta,1} \geq
  (\frac{T^k}{\mathbb E_{\theta,0}[\theta^k]})\eta_{\theta,0}.  \\
\end{equation*}

The other quantity to lower bound is the mean ability of workers who have confidence above $T$, i.e. $\qw {\theta | c \geq T, passed}$. We have 
  \begin{equation*}
    \qw {\theta | c \geq T, passed} \geq T \pr{\theta \geq T|c \geq T,passed}
    \label{}
  \end{equation*}
  and
  \begin{equation*}
    \pr{\theta \geq T|c \geq T,passed}  =\frac{\eta_{\theta,1}\pr{c\geq T|\theta \geq T,passed}}{\eta_{c,1}}.
  \end{equation*}
  Then by equation 13 and 15, we have the desired result. The remaining part of the proof is similar to lemma \ref{lemmaUnsureBer}.\qed
\end{proof}

Denote $B_1 = \frac{T^k}{(\mathbb E_{\theta,0}[\theta^k])^{1-k_0}}, B_2 = \frac{T^k}{\mathbb E_{\theta,0}[\theta^k]}$, the main order term of cost under an unsure mechanism is $\frac{1}{(2B_1T-1)^2B_2\eta_{\theta,0}}$ when $\eta_{c,1} \geq \eta_{\theta,1}$ and $\frac{1}{(2(\mu_{c,1})^{k_1}T-1)^2\eta_{c,1}}$ when $\eta_{c,1} < \eta_{\theta,1}$. As previously discussed, $k$ is usually a small number. Then $B_1,B_2$ scale like constant factors. Thus we can let $B_1 = \frac{(\mu_{\theta,0})^k}{(\mathbb E_{\theta,0}[\theta^k])^{1-k_0}}$ to consider the worst case, and ignore $B_2$. Similar to definition \ref{def2}, we define the $\alpha$-effectiveness for the unsure mechanisms:
\begin{definition}
  % Let $\mathbb E_{\theta,0}[\cdot], \mu_{\theta,0}, \mu_{\theta,1}, \mu_{c,1}, \eta_{\theta,0}, \eta_{c,1}, k, k_0, k_1$ and $T$ be defined as before. Then 
  The unsure mechanism with confidence threshold $T$, utilizing the worker testing stage with $k$ golden standard tasks, is at least {\bf $\alpha$-effective} if 
%  it is provable that for the aggregated label being correct with probability at least $1-\delta$,
  \begin{equation}
    \begin{cases}
      \frac{1}{(2\frac{(\mu_{\theta,0})^k}{(\mathbb E_{\theta,0}[\theta^k])^{1-k_0}}T-1)^2\eta_{\theta,0}} \leq \frac{1}{(2\mu_{\theta,1}-1)^\alpha},\quad \eta_{c,1} \geq \eta_{\theta,1}, \\
      \frac{1}{(2(\mu_{c,1})^{k_1}T-1)^2\eta_{c,1}} \leq \frac{1}{(2\mu_{\theta,1}-1)^\alpha},\quad \eta_{c,1} < \eta_{\theta,1}. \\
    \end{cases}
    \label{equadef3}
  \end{equation}
  \label{def3}
\end{definition}
The $\alpha$-effectiveness again measures the significance of the imporvement on the cost bound, with respect to doing simple aggregation from the crowd after the worker testing stage.
%(1) When $\eta_{c,1} \geq \eta_{\theta,1}$, i.e. the workers tend to be over-confident, the cost bound is still based on $\eta_{\theta,0}$, which depends on the crowd ability distribution $A$. We aim at transforming the analysis to the situation in quality ensured mechanisms and assume equation 14-15 which are the same to equation 7-8. When $\eta_{c,1} < \eta_{\theta,1}$, i.e. on average, the workers who pass the test are under-confident, we focus on the crowd confidence distribution $C$ directly. We introduce the assumptions in equation 17-18 which is similar to the assumptions in equation 7-8. Then we directly analyse $C$. 
Then we can show the condition when an unsure mechanism can be $\alpha$-effective, which is similar to theorem 1. The process of the proof is also similar to theorem 1, thus is omitted.
\begin{theorem}
  %Let $\mathbb E_{\theta,0}[\cdot], \mu_{\theta,0}, \mu_{\theta,1}, \mu_{c,1}, \sigma^2_{\theta,0}, \sigma^2_{c,1}, k, k_0, k_1$ be defined as before. Let $\mathrm{Pr}_{\theta,0}(\cdot)$ be the proability over the crowd ability distribution before the worker testing stage, and $\mathrm{Pr}_{c,1}(\cdot)$ be the proability over the crowd confidence distribution after the worker testing stage. 
  Assume the conditions in equation (1,12-16) to hold. \\
  (1) When $\eta_{c,1} \geq \eta_{\theta,1}$, let 
  \begin{equation}
    m' = (1+\gamma)\frac{(1+\sqrt{1-4\sigma_{\theta,0}^2})^2}{\sqrt{3}\sigma_{\theta,0}^2[2B_1\sigma_{\theta,0}^1+(2B_1\mu_{\theta,0}-1)]^2},\quad B_1 = \frac{(\mu_{\theta,0})^k}{(\mathbb E_{\theta,0}[\theta^k])^{1-k_0}},
  \end{equation}
  \begin{equation}
    \gamma = \frac{\mathrm{Pr}_{\theta,0}(\theta \leq \mu_{\theta,0} - \frac{1}{2}\sqrt{1-\sqrt{1-4\sigma_{\theta,0}^2}})}{\mathrm{Pr}_{\theta,0}(\theta \geq \mu_{\theta,0} + \frac{1}{2}\sqrt{1-\sqrt{1-4\sigma_{\theta,0}^2}})},
  \end{equation}
  and 
  \begin{equation} 
    T= \mu_{\theta,0} + \frac{1}{2}\sqrt{1-\sqrt{1-4\sigma_{\theta,0}^2}}.
  \end{equation}
  (2) When $\eta_{c,1} < \eta_{\theta,1}$, let 
  \begin{equation} 
    m' = (1+\gamma)\frac{(1+\sqrt{1-4\sigma_{c,1}^2})^2}{\sqrt{3}\sigma_{c,1}^2[2\mu_{c,1}^{k_1}\sigma_{c,1}^2+(2\mu_{c,1}^{k_1+1}-1)]^2},
  \end{equation}
  \begin{equation}
    \gamma = \frac{\mathrm{Pr}_{c,1}(c \leq \mu_{c,1} - \frac{1}{2}\sqrt{1-\sqrt{1-4\sigma_{c,1}^2}})}{\mathrm{Pr}_{c,1}(c \geq \mu_{c,1} + \frac{1}{2}\sqrt{1-\sqrt{1-4\sigma_{c,1}^2}})},
  \end{equation}
  and
  \begin{equation} 
    T= \mu_{c,1} + \frac{1}{2}\sqrt{1-\sqrt{1-4\sigma_{c,1}^2}}.
  \end{equation}
 Then if 
 \begin{equation}
   m' \leq \frac{1}{(2\mu_{\theta,1}-1)^\alpha},
   \label{}
 \end{equation}
 the unsure mechanism with confidence threshold $T$, utilizing the working testing stage with $k$ golden standard tasks, is at least $\alpha$-effective.
  \label{theoremUnsureReal}
\end{theorem}
%The above theorem divides into two situations. It can also been seen that if workers tend to be over-confident, i.e. $\eta_{c,1} \geq \eta_{\theta,1}$, then we can focus on estimating the mean and variance of the crowd ability distribution before test, just like the quality ensured mechanism. This can be done by utilizing the feedback on golden standard tasks for test. On the opposite, if workers tend to be under-confident, i.e. $\eta_{c,1} < \eta_{\theta,1}$, then we focus on estimating the mean and variance of the crowd confidence distribution for workers who pass the test. This can be done by asking workers to return their confidence while doing tasks. In applications, the relationship between $\eta_{c,1}$ and $\eta_{\theta,1}$ can be decided via a survey to the crowd. 
%It can also been seen that if workers tend to be over-confident, i.e. $\eta_{c,1} \geq \eta_{\theta,1}$, then we can focus on estimating the mean and variance of the crowd ability distribution before test, just like the quality ensured mechanism. This can be done by utilizing the feedback on golden standard tasks for test. On the opposite, if workers tend to be under-confident, i.e. $\eta_{c,1} < \eta_{\theta,1}$, then we focus on estimating the mean and variance of the crowd confidence distribution for workers who pass the test. This can be done by asking workers to return their confidence while doing tasks. In applications, the relationship between $\eta_{c,1}$ and $\eta_{\theta,1}$ can be decided via a survey to the crowd. 
\section{Discussion on the Payment Strategy}
\label{DPS}
In the above analysis, we assume that returning labels and choosing unsure option are equally paid. In many crowdsourcing applications, this payment strategy may lead to the potential danger for causing workers to always choose the unsure option without returning any labels. This phenomenon violates the assumption that the workers are honest. Using alternative incentive compatible payment method \citep{shah2014double} can be helpful to deal with this problem. As an example, the following payment method incentivizes the workers to behave honestly, under the assumption that the workers aim to maximize their payments: \\
(1) Choosing unsure option is paid for $T$, the value of the confidence threshold. \\
(2) Among the returned labels, the ones that accord with the aggregated label are paid for $1$, otherwise are paid for $0$.\\
It is easy to show that this payment strategy is incentive compatible. If the worker has confidence $c > T$, then the expected payment for returning the label is also $c$, while the payment for choosing unsure option is $T$. Thus the worker is desirable to return the label. If $c < T$ the reason is similar for the worker to choose unsure option. The analysis in previous sections is a good approximation for this payment method. The reason is that, since $T>1/2$, and for an effective unsure mechanism, most of the returned labels should agree with the aggregated label, thus assuming returning labels and choosing unsure option are both paid for $1$ does not sacrifice much tightness for the cost bounds. Overall, it is interesting to study how the optimal incentive compatible payment method and the unsure mechanism can be integrated for different application scenarios. We leave this as future work.

\section{Online Algorithm with Unsure Option}
\label{Alg}
\begin{algorithm}[t]
   \caption{OLU (OnLine algorithm with Unsure option)}
   \label{algUCB}
\begin{algorithmic}
  \STATE {\bfseries Input: crowd $A$, number of rounds $N$, confidence thresholds $t_k, k\in\{1,2,\dots,K\}$}.
   \STATE Random initialize $T_1 \in \{t_1,t_2,\dots,t_K\}$;
   \STATE Initialize $N_j = 0, j\in\{1,2,\dots,K\}$;
   \FOR{$i=1$ {\bfseries to} $N$}
    \STATE Draw one worker and provide unsure option with $T_i$;
    \STATE $r_i =(2T_i-1)^2\mathbb I(c_i \geq T_i)$;
    \FOR{$k=1$ {\bfseries to} $K$}
    \IF{$t_k ==T_i$}
    \STATE $N_k = N_k + 1$;
    \STATE $\hat r_k = \frac{1}{N_k}\big((N_k-1)\hat r_k+r_i\big)$;
    \ENDIF
    \ENDFOR
    \STATE $n = \argmax\limits_{k\in\{1,2,\dots,K\}}(\hat r_k + \sqrt{\frac{2\ln i}{N_k}})$;
    \STATE $T_{i+1} = t_n$;
   \ENDFOR
   \end{algorithmic}
\end{algorithm}

The central task for applying an unsure mechanism is to determine the confidence threshold $T$. According to the previous analysis, setting $T$ can be transfered to the problem of estimating the mean and variance of the crowd ability distribution or the crowd confidence distribution. For the case that we need to consider the crowd ability distribution, doing accurate estimation requires a sufficient number of golden standard tasks with known labels. However, this condition is difficult to be satisfied in practice. For solving this problem, we propose an alternative online bandit based algorithm for setting the confidence threshold $T$. Note that we still allow to use a small number of golden standard tasks to perform the worker testing stage. 

The task is to properly choose the confidence threshold $T$, which can be treated as bandit arms. We can model the crowdsourcing process as the following bandit game: To collect a new label, a random worker is drawn from the crowd, and a confidence threshold $T$ is provided. $T$ is updated online by the bandit algorithm. We consider only discrete candidate set of $t_j,j\in\{1,2,\dots,K\}$, which segments the interval $[0.5,1]$ into finite number of parts. Motivated by the previous analysis, we define the reward as 
\begin{equation}
  r_i = (2T_i-1)^2\mathbb I(c_i \geq T_i),
\label{banditReward}
\end{equation}
in which $i$ denotes the $i$th round, $T_i$ denotes the chosen confidence threshold, and $\mathbb I(c_i \geq T)$ denotes the indicator function of the event that the worker does not choose the unsure option. Under this definition, for each arm $t_k$, let $N_k$ denote the number of times the arm is chosen. The average reward $\hat r_k = \frac{1}{N_k}\sum_{n=1}^{N_k}(2t_k-1)^2\mathbb I(c_i \geq t_k)$ is the empirical estimation of $(2t_k-1)^2\pr{c_i\geq t_k}$, i.e., the inverse of main order term of cost for accurate estimation, which should be maximized. $r_i$ are $i.i.d.$ random variables bounded in $[0,1]$. The above problem can be solved by many bandit optimization methods, such as the UCB-1 algorithm \citep{auer2002finite}, which is illustrated in algorithm \ref{algUCB}. Note that more sophisticated bandit algorithm can be designed in this task, since the sample collected on one arm may provide additional information on other arms. As the major topic in this paper is theoretical analsys other than algorithm design, we leave this as future work.
\begin{figure*}[t]
  \centering
\includegraphics[width=.95\textwidth]{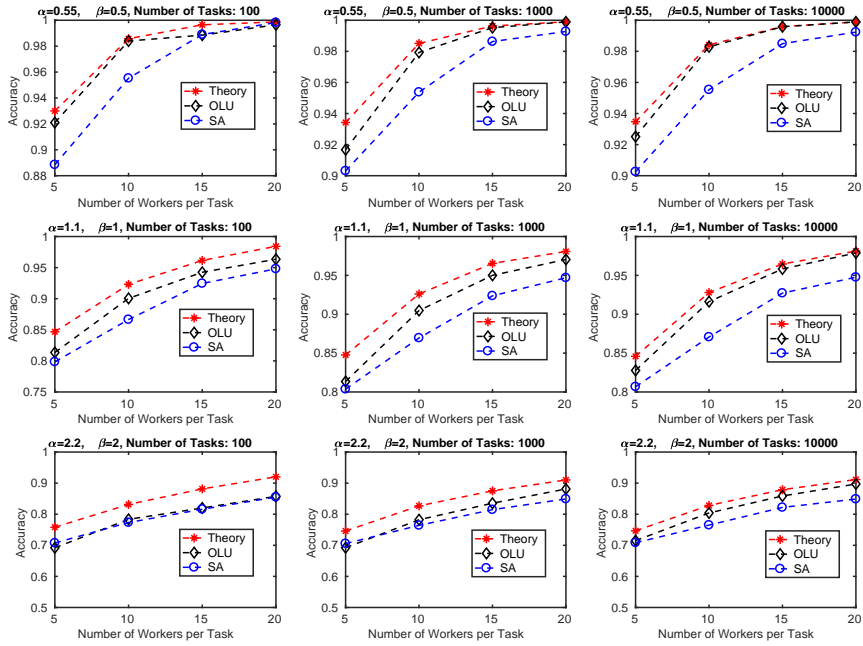}
\caption{Experimental results. In the legend, ``Theory'' denotes the result from setting the confidence threshold as equation 25. ``OLU'' denotes the results from online unsure mechanism algorithm. ``SA'' denotes simple aggregation without unsure mechanism.}
\label{figExp}
\end{figure*}

\section{Experiments}
\label{Exp}
We used synthetic data to test the theoretical results and the online algorithm. In the experiments, a set of binary labeling tasks were generated, and the ground-truth labels were uniformly sampled from $\{-1,1\}$. To simulate on different types of crowd ability distributions, the abilities of the workers were sampled from different Beta distributions. The choices of distribution parameters $\{\alpha,\beta\}$ were $\{0.55,0.5\}, \{1.1,1\}$ and $\{2.2,2\}$. The corresponding mean was $0.52$ and corresponding variances were $0.1217, 0.0805$ and $0.0480$. Since the mean is close to $1/2$, the left hand side of the distributions were not cut according to equation 1. The returned labels were sampled from the Bernoulli distributions according to the abilities. To simulate the situation that the confidence can be largely deviated from the ability, we assumed that the unsure option was used when the sampled ability is above $T$ or below $1-T$.

For each task, we collected the same number of returned labels for majority voting aggregation. The baseline method was simple aggregation without using the unsure mechanism. For examining theory, we adopted the unsure mechanism, on which the confidence threshold is set according to equation 23. To implement the online algorithm, we employed the candidate threshold set $\{0.55,0.60,\dots,1\}$. For the fairness of the comparison, we adopted the worker testing stage with $k=1$ for all methods. 

The results are illustrated in Figure \ref{figExp}. In all kinds of crowd ability distributions, employing unsure mechanism outperformed simple aggregation. As the variance got larger, the number of workers needed for high accuracy aggregation was significantly reduced. Furthermore, when the number of tasks got larger, the performance of the online algorithm got better. This phenomenon indicates that the online algorithm is capable to be employed when the number of tasks to be done is large.
\section{Conclusions} 
\label{Conc}
In this work, we theoretically study the cost-saving effect of the crowdsourcing with unsure option setting. We give the sufficient condition for an unsure mechnanism can lead to significant cost reduction, show how confidence threshhold can be properly set. Motivated by the theoretical analysis, we also propose an alternative online algorithm for setting the confidence threshold. We also hope our work to be a motivation for further studies on how crowdsourcing can be helped by utilizing subjective uncertainty of workers.

\bibliography{example_paper}

\begin{thebibliography}{25}
\providecommand{\natexlab}[1]{#1}
\providecommand{\url}[1]{{#1}}
\providecommand{\urlprefix}{URL }
\expandafter\ifx\csname urlstyle\endcsname\relax
  \providecommand{\doi}[1]{DOI~\discretionary{}{}{}#1}\else
  \providecommand{\doi}{DOI~\discretionary{}{}{}\begingroup
  \urlstyle{rm}\Url}\fi
\providecommand{\eprint}[2][]{\url{#2}}

\bibitem[{Abbasi-Yadkori et~al(2015)Abbasi-Yadkori, Bartlett, Xi, and
  Malek}]{abbasilarge}
Abbasi-Yadkori Y, Bartlett PL, Xi C, Malek A (2015) Large-scale markov decision
  problems with kl control cost and its application to crowdsourcing. In:
  Proceedings of the 32th International Conference on Machine Learning

\bibitem[{Abraham et~al(2013)Abraham, Alonso, Kandylas, and
  Slivkins}]{abraham2013adaptive}
Abraham I, Alonso O, Kandylas V, Slivkins A (2013) Adaptive crowdsourcing
  algorithms for the bandit survey problem. Proceedings of the 26th Conference
  on Learning Theory

\bibitem[{Auer et~al(2002)Auer, Cesa-Bianchi, and Fischer}]{auer2002finite}
Auer P, Cesa-Bianchi N, Fischer P (2002) Finite-time analysis of the multiarmed
  bandit problem. Machine Learning 47(2):235--256

\bibitem[{Boucheron et~al(2013)Boucheron, Lugosi, and
  Massart}]{Boucheron2013Concerntration}
Boucheron S, Lugosi G, Massart P (2013) Concentration inequalities: A
  nonasymptotic theory of independence. Oxford university press

\bibitem[{Chen et~al(2013)Chen, Lin, and Zhou}]{chen2013optimistic}
Chen X, Lin Q, Zhou D (2013) Optimistic knowledge gradient policy for optimal
  budget allocation in crowdsourcing. In: Proceedings of the 30th International
  Conference on Machine Learning, pp 64--72

\bibitem[{Dalvi et~al(2013)Dalvi, Dasgupta, Kumar, and
  Rastogi}]{dalvi2013aggregating}
Dalvi N, Dasgupta A, Kumar R, Rastogi V (2013) Aggregating crowdsourced binary
  ratings. In: Proceedings of the 22nd International Conference on World Wide
  Web, pp 285--294

\bibitem[{Dawid and Skene(1979)}]{dawid1979maximum}
Dawid AP, Skene AM (1979) Maximum likelihood estimation of observer error-rates
  using the em algorithm. Applied statistics pp 20--28

\bibitem[{Dekel and Shamir(2009{\natexlab{a}})}]{dekel2009good}
Dekel O, Shamir O (2009{\natexlab{a}}) Good learners for evil teachers. In:
  Proceedings of the 26th Annual International Conference on Machine Learning,
  pp 233--240

\bibitem[{Dekel and Shamir(2009{\natexlab{b}})}]{dekel2009vox}
Dekel O, Shamir O (2009{\natexlab{b}}) Vox {P}opuli: collecting high-quality
  labels from a crowd. In: Proceedings of the 20nd Annual Conference on
  Learning Theory

\bibitem[{Ho et~al(2013)Ho, Jabbari, and Vaughan}]{ho2013adaptive}
Ho CJ, Jabbari S, Vaughan JW (2013) Adaptive task assignment for crowdsourced
  classification. In: Proceedings of the 30th International Conference on
  Machine Learning, pp 534--542

\bibitem[{Jain et~al(2014)Jain, Gujar, Bhat, Zoeter, and
  Narahari}]{jain2014incentive}
Jain S, Gujar S, Bhat S, Zoeter O, Narahari Y (2014) An incentive compatible
  multi-armed-bandit crowdsourcing mechanism with quality assurance. arXiv
  preprint arXiv:14067157

\bibitem[{Karger et~al(2011)Karger, Oh, and Shah}]{karger2011iterative}
Karger DR, Oh S, Shah D (2011) Iterative learning for reliable crowdsourcing
  systems. In: Advances in Neural Information Processing Systems 24, pp
  1953--1961

\bibitem[{Li and Yu(2014)}]{li2014error}
Li H, Yu B (2014) Error rate bounds and iterative weighted majority voting for
  crowdsourcing. arXiv preprint arXiv:14114086

\bibitem[{Raykar et~al(2010)Raykar, Yu, Zhao, Valadez, Florin, Bogoni, and
  Moy}]{raykar2010learning}
Raykar VC, Yu S, Zhao LH, Valadez GH, Florin C, Bogoni L, Moy L (2010) Learning
  from crowds. Journal of Machine Learning Research 11:1297--1322

\bibitem[{Shah and Zhou(2015)}]{shah2014double}
Shah NB, Zhou D (2015) Double or nothing: Multiplicative incentive mechanisms
  for crowdsourcing. In: Advances in Neural Information Processing Systems 28,
  pp 1--9

\bibitem[{Shalev-Shwartz and Ben-David(2014)}]{shalev2014understanding}
Shalev-Shwartz S, Ben-David S (2014) Understanding machine learning: From
  theory to algorithms. Cambridge University Press

\bibitem[{Tran-Thanh et~al(2013)Tran-Thanh, Venanzi, Rogers, and
  Jennings}]{tran2013efficient}
Tran-Thanh L, Venanzi M, Rogers A, Jennings NR (2013) Efficient budget
  allocation with accuracy guarantees for crowdsourcing classification tasks.
  In: Proceedings of the 2013 International Conference on Autonomous Agents and
  Multi-agent Systems, pp 901--908

\bibitem[{Urner et~al(2012)Urner, Ben-David, and Shamir}]{urner2012learning}
Urner R, Ben-David S, Shamir O (2012) Learning from weak teachers. In:
  Proceedings of 15th International Conference on Artificial Intelligence and
  Statistics, pp 1252--1260

\bibitem[{Wang and Zhou(2016)}]{wang2016cost}
Wang L, Zhou ZH (2016) Cost-saving effect of crowdsourcing learning. In:
  Proceedings of the 25th International Joint Conference on Artificial
  Intelligence

\bibitem[{Wang and Zhou(2015)}]{wang2015crowdsourcing}
Wang W, Zhou ZH (2015) Crowdsourcing label quality: a theoretical analysis.
  Science China Information Sciences 58(11):1--12

\bibitem[{Zhang et~al(2014)Zhang, Chen, Zhou, and Jordan}]{zhang2014spectral}
Zhang Y, Chen X, Zhou D, Jordan MI (2014) Spectral methods meet {EM}: a
  provably optimal algorithm for crowdsourcing. In: Advances in Neural
  Information Processing Systems 27, pp 1260--1268

\bibitem[{Zhong et~al(2015)Zhong, Tang, and Zhou}]{zhongactive}
Zhong J, Tang K, Zhou ZH (2015) Active learning from crowds with unsure option.
  In: Proceedings of the 24th International Joint Conference on Artificial
  Intelligence

\bibitem[{Zhou et~al(2012)Zhou, Basu, Mao, and Platt}]{zhou2012learning}
Zhou D, Basu S, Mao Y, Platt JC (2012) Learning from the wisdom of crowds by
  minimax entropy. In: Advances in Neural Information Processing Systems 25, pp
  2195--2203

\bibitem[{Zhou et~al(2015)Zhou, Liu, Platt, Meek, and
  Shah}]{zhou2015regularized}
Zhou D, Liu Q, Platt JC, Meek C, Shah NB (2015) Regularized minimax conditional
  entropy for crowdsourcing. arXiv preprint arXiv:150307240

\bibitem[{Zhou et~al(2014)Zhou, Chen, and Li}]{zhou2014optimal}
Zhou Y, Chen X, Li J (2014) Optimal pac multiple arm identification with
  applications to crowdsourcing. In: Proceedings of the 31st International
  Conference on Machine Learning, pp 217--225

\end{thebibliography}
\bibliographystyle{spbasic}      % basic style, author-year citations
\end{document}